\def\year{2022}\relax
\documentclass[letterpaper]{article} 
\usepackage{aaai22}  
\usepackage{times}  
\usepackage{helvet}  
\usepackage{courier}  
\usepackage[hyphens]{url}  
\usepackage{graphicx} 
\usepackage{natbib}  
\usepackage{caption} 
\DeclareCaptionStyle{ruled}{labelfont=normalfont,labelsep=colon,strut=off} 
\usepackage{multirow}
\usepackage{graphicx}

\usepackage{algorithm}
\usepackage{algorithmic}
\usepackage{amsmath}
\usepackage{booktabs}
\usepackage{xcolor}
\usepackage{soul}
\newcommand*{\Scale}[2][4]{\scalebox{#1}{$#2$}}%

\definecolor{neonfuchsia}{rgb}{1.0, 0.25, 0.39}

\usepackage[hang,flushmargin]{footmisc}
\newcommand\workshopnote[1]{\renewcommand\thefootnote{}\footnote{#1}}
 
\usepackage{newfloat}
\usepackage{listings}
\floatstyle{ruled}
\newfloat{listing}{tb}{lst}{}
\floatname{listing}{Listing}
\title{\textit{KAER}: A Knowledge Augmented Pre-Trained Language Model for Entity Resolution}

\author {
    Liri Fang\textsuperscript{\rm 1\equalcontrib},
    Lan Li\textsuperscript{\rm 1\equalcontrib},
    Yiren Liu\textsuperscript{\rm 2},
    Vetle I. Torvik\textsuperscript{\rm 1},
    Bertram Lud\"ascher\textsuperscript{\rm 1}
}
\affiliations {
    \textsuperscript{\rm 1} School of Information Sciences, University of Illinois Urbana-Champaign\\
    \textsuperscript{\rm 2} Informatics, University of Illinois Urbana-Champaign\\
    lirif2@illinois.edu, lanl2@illinois.edu, yirenl2@illinois.edu, vtorvik@illinois.edu, ludaesch@illinois.edu
}

\usepackage{bibentry}

\begin{document}

\maketitle

\begin{abstract}

Entity resolution has been an essential and well-studied task in data cleaning research for decades. 
Existing work has discussed the feasibility of utilizing pre-trained language models to perform entity resolution and achieved promising results.
However, few works have discussed injecting domain knowledge to improve the performance of pre-trained language models on entity resolution tasks. 
In this study, we propose \textbf{K}nowledge \textbf{A}ugmented \textbf{E}ntity \textbf{R}esolution (\textit{KAER}), a novel framework named for augmenting pre-trained language models with external knowledge for entity resolution. We discuss the results of utilizing different knowledge augmentation and prompting methods to improve entity resolution performance. 
Our model improves on Ditto, the existing state-of-the-art entity resolution method. In particular, 1) \textit{KAER} performs more robustly and achieves better results on ``dirty data'', and 2) with more general knowledge injection, \textit{KAER} outperforms the existing baseline models on the textual dataset and dataset from the online product domain. 3) \textit{KAER} achieves competitive results on highly domain-specific datasets, such as citation datasets, requiring the injection of expert knowledge in future work. 
\vspace{-0.86cm}
\end{abstract}
\workshopnote{Accepted to Workshop on Knowledge Augmented Methods for Natural Language Processing, in conjunction with AAAI 2023.}

\section{Introduction}
Recent studies using \emph{transformer-based Pre-trained Language Models} (PLMs) have shown their strong ability to perform various types of NLP tasks \cite{min_recent_2021}. However, few studies have discussed the application of PLMs in the domain of data cleaning~\cite{li_deep_2020,narayan_can_2022,vos2022towards}.  
Entity resolution is a common data cleaning task that aims to identify the entries referring to the same real-world entities within or across databases~\cite{christen_data_2012}. 




Most existing techniques on entity resolution assume the same schema for records from different sources \cite{elmagarmid_duplicate_2007}. However, in many situations, raw records are obtained from heterogeneous sources and use different schema \cite{enriquez_entity_2017, arabnia_when_2021}. In addition, source data is often from varied domains (e.g., publications, online products, musicians) and in different formats (e.g., numerical, textual, geolocations). 
All of these increase the difficulty for practitioners to perform entity resolution tasks without prior knowledge of the domain-specific information about the data.
Thus, we hypothesize that enhancing the external knowledge at the schema and entity level can improve entity resolution tasks.  

With transformer-based PLMs, recent studies draw increasing attention to entity resolution problems~\cite{li_deep_2020, trabelsi_dame_2022}. However, current studies show that the performance might not be ideal when simply inputting the serialized entity pairs into PLMs for classification. Ditto \cite{li_deep_2020} injects domain information: pre-defined entity types (i.e., PRODUCT and NUM), and standardizes the numerical formats to improve the performance before feeding the serialized entity pairs into PLMs.

We push this idea further by injecting more external knowledge at the schema and entity level. Knowledge injection at the schema level aims to infer the fine-grained semantic types (e.g., ALBUM, ARTIST, PUBLISHER) for each column based on data values. 
For the entity level, entity mentions are identified from WikiData and annotated in the initial text with semantic type information of the linked entities. In addition, different formats used to inject external knowledge into the initial entity pairs may vary the performance of PLMs. Thus, three prompting methods are further explored in this study: space, slash, and constrained tuning of PLMs.

To summarize, starting from state-of-the-art method Ditto \cite{li_deep_2020}, we propose a framework for  \textbf{K}nowledge \textbf{A}ugmented \textbf{E}ntity \textbf{R}esolution (KAER):
\begin{itemize}
    \item using \textbf{C}olumn \textbf{S}emantic \textbf{T}ype (CST) inference and \textbf{E}ntity \textbf{L}inking (EL) in order to inject domain-specific information as additional signals to pre-trained language models. 
    \item leveraging three prompting methods to better augment the acquired knowledge to PLMs.
    \item analyzing the effectiveness of different combinations of knowledge injection and prompting methods on entity resolution tasks from different domains and data types.
\end{itemize}

    

\section{Related Work}
\begin{figure*}[!ht]
    \centering
    \includegraphics[width=\linewidth]{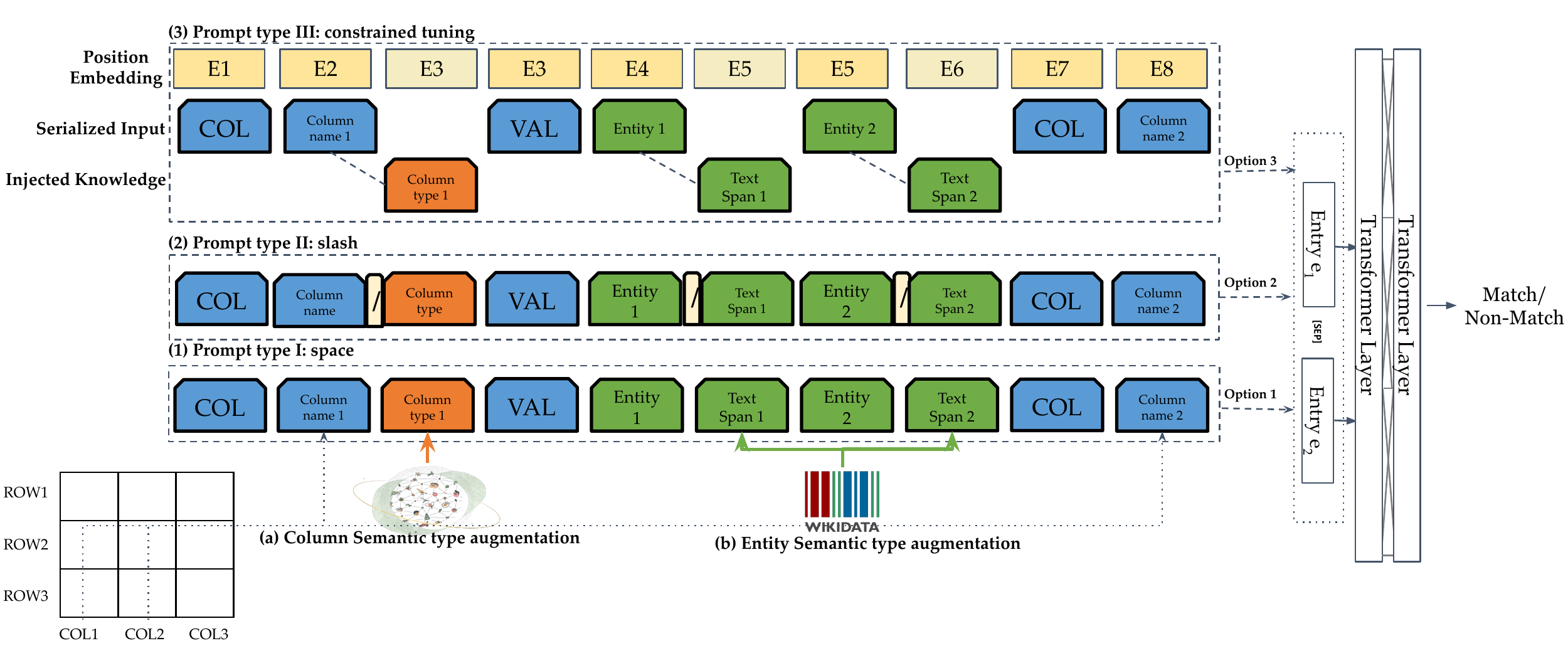}
    \vspace{-0.7cm}
    \caption{The framework of KAER. First, (a) column semantic type augmentation and (b) entity semantic type augmentation are used to inject domain knowledge from both column and entity levels of the input table. 
    Then three options of prompting types are provided after external knowledge injection: (1) using space; (2) using ``/''; (3) using constrained tuning. 
    With the knowledge injection and prompting methods, the serialized sequence is fed into the PLMs.}
    \vspace{-0.5cm}
    \label{fig:framework}
\end{figure*}

\subsection{Pretrained Model for Entity Resolution}


A few recent works apply transformer-based PLMs to entity resolution tasks. \citet{paganelli_analyzing_2022} discover that simply fine-tuning BERT can benefit matching/non-matching classification tasks and recognize the input sequence as a pair of records. \citet{li_improving_2021} leverage siamese network structure with PLMs to improve the efficiency of PLMs during the blocking phase. 
Ditto by \cite{li_deep_2020} is the state-of-the-art entity matching system based on PLMs, i.e., RoBERTa. In addition, Ditto provides a deeper language understanding for entity resolution by injecting domain knowledge, summarizing the key information, and augmenting with more difficult examples for training data. 
 
\vspace{-0.5em}
\subsection{Knowledge Augmentation}
Recent studies on semantic column type augmentation and entity linking can inject external domain knowledge into PLMs at the schema and entity levels. 
Semantic column-type augmentation can inject domain-specific knowledge for columns with/without existing column names. Existing studies~\cite{hulsebos_sherlock_2019,suhara_annotating_2022} use deep learning approaches to detect semantic data types at the column level.
Entity linking \cite{li_deep_2020} refers to linking entity mentions appearing in natural language text with their corresponding entities in an external knowledge base, e.g., Wikidata. 
\citet{ayoola_refined_2022} introduces an entity linking method by fine-tuning a PLM over Wikidata, which is used for entity linking in this study. 

``prompting'' is the method of conditioning the language model~\citet{pengfeiliu}. \citet{kojima_large_2022} propose Zero-shot-CoT, a single zero-shot prompt that highlights that the performance of the language model has been affected by different templates of the prompt.
Furthermore, the method to inject the identified external knowledge into PLMs matters. \citet{liu_k-bert_2020} propose the model K-BERT adding soft-position encoding and visible matrix to the augmented input sequence to reduce knowledge noise injected into the original input and corresponding models. 

\section{Methodology}

\subsection{Notation of Entity Resolution Task}
The formulation of the entity resolution task in this paper is based on the current work by~\citet{li_deep_2020}. The input of the entity resolution task consists of a set $M \subseteq D_1 \times D_2$, where $D_1$ and $D_2$ are two sets of data entry collections that contain duplicated entries.  
For each data entry, $e \in {(col_i, val_i)}_{1 \leq i \leq N}$ where $N$ is the number of columns, and $(e_1, e_2) \in M$. The task discussed in this paper focuses on: for each data entry pair $(e_1, e_2) \in M$, determine whether $e_1$ and $e_2$ refer to the same data entity.

\subsection{Model Overview}
This study introduces a new framework for entity resolution: KAER (See Figure~\ref{fig:framework}). 
KAER uses PLMs for entity resolution and contains three modules for knowledge augmentation: a) column semantic type augmentation, b) entity semantic type augmentation, and c) three options of prompting types. 
The following sections first describe the PLMs for entity resolution and then each module.

\subsubsection{Pre-trained language model for entity resolution.}
Following the work by~\citet{li_deep_2020}, KAER uses RoBERTa as the backbone model. For each data entry pair, $(e_1, e_2)$, 
 the text context of column names and values of $e_1$ and $e_2$ are serialized and concatenated as the input for PLMs. The [CLS] token position is used to classify whether $e_1$ and $e_2$ refer to the same entity. 
The loss for optimizing the classification objective is:
\begin{equation}
    \ell = - log\ p(y | s(e_1, e_2))
\end{equation}

where y denotes whether $e_1$ and $e_2$ refer to the same entity, and $s(\cdot,\cdot)$ denotes the serialization and transformation of entity pairs with knowledge injection and prompting methods. 
\begin{equation}
\Scale[0.8]{
    s(e_i, e_j) ::=  \text{[CLS]}\ serialize(e_i)\ \text{[SEP]}\ serialize(e_j)\ \text{[SEP]}} 
\end{equation}
where $serialize(\cdot)$ serializes each data entry.
\begin{equation}\Scale[0.8]{
\begin{split}
serialize(e_i) ::= & \text{[COL]}\ f(col_1,pt)\ \text{[VAL]}\ g(val_1,pt)\ ...\\
            & \text{[COL]}\ f(col_N,pt)\ \text{[VAL]}\ g(val_N,pt)   
\end{split}}
\end{equation}
where $f(col_i, pt)$ denotes the semantic column type injection with prompting method $pt$, and $g(val_i, pt)$ denotes the entity linking injection with prompting method $pt$. The following sections describe  column semantic type injection, entity linking, and three prompting methods.


\subsubsection{Column semantic type augmentation.}
Column type prediction aims to predict the semantic type of each column by considering the intra-column context, so that fine-grained semantic information can be augmented to the framework. For example, ``name'' as a column name can provide more semantic information if injected with the semantic type ``song''.  
Here, KAER uses Sherlock~\cite{hulsebos_sherlock_2019} for column semantic type prediction. In detail, \textit{Sherlock} is a multi-input deep neural network pre-trained on 686,765 data columns retrieved from the 
VizNet corpus by matching 78 semantic data types from DBpedia to column headers \cite{hulsebos_sherlock_2019}. 




\begin{table*}[!h]
\centering
\resizebox{\linewidth}{!}{
\begin{tabular}{l|ll|llll|l}
\hline
\multicolumn{1}{l|}{}                  & \multicolumn{2}{c|}{\textbf{Dirty}} & \multicolumn{4}{c|}{\textbf{Structured}}                      & \textbf{Textual} \\ \hline
\multicolumn{1}{l|}{\textbf{Models}} &
  \textbf{\begin{tabular}[l]{@{}l@{}}DBLP\\ -\\ GoogleScholar\end{tabular}} &
  \multicolumn{1}{c|}{\textbf{\begin{tabular}[l]{@{}l@{}}iTunes\\ -\\ Amazon\end{tabular}}} &
  \textbf{\begin{tabular}[l]{@{}l@{}}iTunes\\ -\\ Amazon\end{tabular}} &
  \textbf{\begin{tabular}[l]{@{}l@{}}Amazon\\ -\\ Google\end{tabular}} &
  \textbf{\begin{tabular}[l]{@{}l@{}}DBLP\\ -\\ ACM\end{tabular}} &
  \multicolumn{1}{l|}{\textbf{\begin{tabular}[l]{@{}l@{}}DBLP\\ -\\ GoogleScholar\end{tabular}}} &
  \textbf{\begin{tabular}[c]{@{}l@{}}Abt\\ -\\ Buy\end{tabular}} \\ \hline
RoBERTa                          & 95.81                             & 72.73                       & 61.76                       & 73.12                       & 98.77                   & 95.32                             & 88.41                                \\
Ditto                            & 95.54                             & 60.38                       & 47.06                       & 74.44                       & 98.55                   & 95.85                             & 88.89                                \\ \hline
KAER ( Roberta + CST )           & 95.47                             & 62.96                       & 80.70**++                   & 68.70                       & 98.77                   & 95.05                             & 92.27**++                            \\
KAER ( Roberta + CST + / )       & 95.74                             & 54.90*                      & \textbf{89.66**++}          & 74.85                       & \textbf{98.99}          & 95.14                             & \textbf{92.61**++}                   \\
KAER ( Roberta + CST + PCT)      & 95.20                             & 72.00                       & 57.14                       & \textbf{76.25}              & 98.09*                  & 95.62                             & 78.24**++                            \\
KAER ( Roberta + EL )            & 95.12                             & 50.00*                      & 61.54                       & 73.77                       & 98.54                   & 95.72                             & 89.11                                \\
KAER ( Roberta + EL + / )        & 95.75                             & \textbf{81.82}              & 86.21**++                   & 74.02                       & 98.43                   & \textbf{96.11*}                   & 88.19                                \\
KAER ( Roberta + EL + PCT)       & 94.80**+                          & 77.78+                      & 67.69                       & 75.63                       & 97.04**+                & 95.29                             & 73.80**++                            \\
KAER ( Roberta + CST + EL )      & \textbf{95.87}                    & 47.06*                      & 59.26                       & 72.28                       & 98.43                   & 95.65                             & 89.66                                \\
KAER ( Roberta + CST + EL + / )  & 95.63                             & 63.33*                      & 57.14+                      & 75.15                       & 98.34                   & 95.67                             & 89.81                                \\
KAER ( Roberta + CST + EL + PCT) & 94.24**++                         & 71.11+                      & 61.29                       & 71.37*                      & 96.26**++               & 94.93+                            & 67.20**++                            \\ \hline
\end{tabular}
}
\vspace{-0.2cm}
\caption{Experimental results using different knowledge injection methods, measured by F1 score. 
The two baseline models: RoBERTa (without domain knowledge injection); and Ditto (RoBERTa and Ditto's general domain knowledge injection methods). 
Other models represent KAER with various knowledge injection and prompting methods combined with RoBERTa. ``+CST'' indicates \textbf{C}olumn \textbf{S}emantic \textbf{T}ype injection.``+EL'' indicates knowledge injection with \textbf{E}ntity \textbf{L}inking. ``+/'' represents prompting with a slash. ``+PCT'' represents \textbf{P}rompting with \textbf{C}onstrained \textbf{T}uning, i.e., soft positions and visible matrix. The paired t-test is processed on between KAER injections and two baseline models, i.e., RoBERTa(*) and Ditto(+). \textbf{**} and \textbf{++} represent 99\% confidence level, and \textbf{*} and \textbf{+} represent 95\% confidence level.}
\label{tab:injection_results}
\vspace{-0.5cm}
\end{table*}

\subsubsection{Entity semantic type augmentation.}
Entity semantic type augmentation leverages the entity linking method to identify all entity mentions from a given knowledge base (KB) within the target text sequence. 
More specifically, 1) to identify the text span of each entity mention $m \in M$, and 2) map each mention to the entity set from an external KB $\mathcal{M}: M \rightarrow E$. 
Thus, additional knowledge from the KB is used to augment the input text records.
In this study, WikiData is used as the external KB since it covers a wide range of domains and is suitable for entity resolution over different datasets. 
KAER uses the state-of-the-art method proposed by~\citet{ayoola_refined_2022}, which uses a RoBERTa model jointly pre-trained over entity typing and entity description modeling objectives to inject coarse entity types. 

\subsubsection{Prompting methods}
\paragraph{Template-based prompting.}
After acquiring domain knowledge, we experiment using text-based templates to combine the knowledge with the initial text input. 
Two different characters are used to concatenate the original entity mention/column name, i.e. slash (``/'') and space. 
The slash symbol is chosen because it contains a semantic meaning equivalent to 'or' in general web text.


\paragraph{Prompting with constrained tuning}
Inspired by hard-soft prompt hybrid tuning~\cite{han_2021},  prompting with constrained tuning is used to generate the token embedding by controlling the visible area of the augmented knowledge only to its corresponding entity mentions or column names. This type of domain knowledge injection modifies the initial input sequence in two dimensions, including the token position and token semantics. More specifically, the token from the initial input sequence has a different absolute position when the template-based prompting is concatenated. In addition, the injected knowledge for a specific entity mention is shared with the surrounding tokens in the same sequence. These two modifications might introduce erroneous knowledge and overload to the framework.

To limit the erroneous knowledge and overload, we leverage constrained tuning, i.e., soft-position encoding and visible matrix, as an additional input of our backbone model, inspired by K\-Bert~\cite{liu_k-bert_2020}. Here, the knowledge-injected sequence is constructed into a tree structure, where the injected knowledge subsequences are considered branches. In this way, as shown in Figure~\ref{fig:framework}, the soft-position encoding assigns the initial tokens the same position encoding in the knowledge-injected sequence as those in the initial input sequence, and assigns the injected knowledge tokens the position in the branches. The visible matrix $V\in \mathcal{R}^{N \times N}$ is a binary matrix, where the injected knowledge token is only visible to itself and the corresponding entity mention or column name tokens, and set as 1, denoted as follows:
\begin{equation}
    V_{ij} = \begin{cases}
    1, & \text{if $token_i$ \text{and} $token_j$ co-occur}.\\
    0, & \text{otherwise}.
  \end{cases}
  \label{Eq:VM}
\end{equation}
where the co-occurrence of $token_i$ and $token_j$ indicates that the tokens co-occur in the initial input sequences or co-occur as head and tail in domain knowledge injection. Here, the head means the entity mention/column name, and the tail indicates the injected knowledge text.

\section{Experiments}
\subsection{Dataset}

KAER is evaluated on the Magellan datasets \cite{magellandata} across various domains. 
The overview of the datasets used in this study is shown in Table~\ref{tab:dataset_overview}. 
The Magellan datasets contain three types of datasets: dirty dataset, structured dataset, and textual dataset.
The dirty dataset is generated from the structured dataset by randomly removing attribute values and appending the initial values to another randomly selected attribute~\cite{li_deep_2020}.

\begin{table}[h]
\centering
\resizebox{\columnwidth}{!}{%
\begin{tabular}{lclccc}
\hline
\textbf{Data Type} & \textbf{Dataset} & \textbf{Domain} & \textbf{Size} & \textbf{\# Positive} & \textbf{\# Attr.} \\ \hline
Structured             & Amazon-Google      & software & 11,460 & 1,167 & 3 \\
                       & iTunes-Amazon      & music    & 539    & 132   & 8 \\
                       & DBLP-ACM           & citation & 12,363 & 2,220 & 4 \\
                       & DBLP-GoogleScholar & citation & 28,707 & 5,347 & 4 \\ \hline
\multirow{3}{*}{Dirty} & iTunes-Amazon      & music    & 539    & 132   & 8 \\
                       & DBLP-ACM           & citation & 12,363 & 2,220 & 4 \\
                       & DBLP-GoogleScholar & citation & 28,707 & 5,347 & 4 \\ \hline
Textual                & Abt-Buy            & product  & 9,575  & 1,028 & 3 \\ \hline
\end{tabular}
}
\vspace{-0.2cm}
\caption{Dataset Summary}
\label{tab:dataset_overview}
\vspace{-0.8cm}
\end{table}

\subsection{Results}
The experimental results are presented in Table \ref{tab:injection_results}.

\textit{Knowledge augmentation performs better on smaller datasets.}
According to the experimental results, incorporating external knowledge can improve the performance on entity resolution tasks. In particular, the proposed knowledge augmentation methods outperform the two baseline models in a data-scarce context.
For example, \textbf{KAER (RoBERTa+EL+/)} outperforms both baseline models Roberta and Ditto by 9.09\% and 21.44\% $\uparrow$ on dataset Dirty/iTunes-Amazon, but not with statistic significance. Meanwhile, it statistically significantly outperforms both baseline models 24.45\% and 39.15\% $\uparrow$ on the dataset Structured/iTunes-Amazon. The two baseline models perform worse on smaller datasets because the language model might not be able to learn enough information to distinguish between different entities when fine tuning, given fewer training instances correctly. 
Prompting with PCT achieves better performance in datasets with smaller text lengths. For example, \textbf{KAER (RoBERTa+CST+PCT)} reaches the best F1 score (76.25\%) without statistic significance on dataset Structured/Amazon-Google, which contains the smallest length of serialized data entry. Moreover, \textbf{KAER (RoBERTa+CST+PCT)} on Structured/Amazon-Google makes a significant difference compared to both baseline models, at 99\% confidence level.

\textit{Knowledge augmentation does not perform well on datasets from certain domains.}
The datasets used in our experiments span several different domains. Using our proposed knowledge augmentation methods (CST and EL), records from certain domains, such as music and online product, benefit from more accurate retrieved knowledge.
For example, \textbf{KAER (RoBERTa+CST+/)} achieves the best F1-score with statistic significance, i.e., 92.61\%, on the textual dataset Abt-Buy from the online product domain.
\textbf{KAER (RoBERTa+EL+/)} achieves the best F1-score (81.82\%) without statistical significance on the musical dataset Dirty/iTunes-Amazon. 
One potential reason for this notable improvement is that the entities mentioned in the datasets from these domains are more general.
However, publication datasets from domains like citation require more domain-specific or even expert knowledge rather than general commonsense. Our knowledge injection methods are primarily designed for retrieving general knowledge.
Thus, for datasets, like Structured/DBLP-ACM, \textbf{KAER (RoBERTa+CST)} performs as well as Roberta and significantly makes a difference with Ditto at 95\% confidence level, and \textbf{KAER (RoBERTa+CST+/)}  achieves the best F1-score 98.99\% without statistic significance. But all the other KAER models perform worse. For future improvement, additional external expert knowledge from the citation domain should be incorporated for knowledge injection. 

\textit{Knowledge augmentation is affected by data quality.}
KAER, with certain knowledge augmentation methods, outperforms the baseline models on dirty datasets. For instance, \textbf{KAER (RoBERTa+CST+EL)} slightly exceeds RoBERTa, and Ditto on Dirty/DBLP-GoogleScholar.
However, the F1 scores of all the other injections on the Dirty/DBLP-GoogleScholar cannot compete with the results on the baseline models (RoBERTa and Ditto). One reason might be that the misleading predictions based on the dirty input by CST result in semantic noise and propagate to the PLMs.
On the other hand, knowledge augmentation with PCT can advance the robustness of the model dealing with low-quality injections. For instance, \textbf{KAER (RoBERTa+CST+PCT)} obtains better F1 score than \textbf{KAER (RoBERTa+CST)} (72.00\% vs. 62.96\%). 
Similar improvement applies to \textbf{KAER (RoBERTa+EL+PCT)} vs. \textbf{KAER (RoBERTa+EL)} 
and  \textbf{KAER (RoBERTa+CST+EL+PCT)} vs. \textbf{KAER (RoBERTa+CST+EL)}. 

\vspace{-0.2cm}



\section{Conclusion}

In this study, we introduced a novel framework, KAER, for entity resolution problems by augmenting PLMS with external knowledge using prompting techniques. 
The framework is evaluated over a set of datasets from varied domains and data types (i.e., dirty, structured, and textual) and improves performance in certain applications over the baseline models. 
Further analysis revealed the effectiveness of different domain knowledge injection and prompting methods on various datasets from different domains and data types.
The results show a promising direction of injecting knowledge to improve entity resolution performance using PLMs. 
Future studies can improve upon our method by introducing more domain-specific knowledge. 

\section{Acknowledgement}
The experiments are conducted on the NCSA Delta supercomputing system, supported by the National Science Foundation under Grant No. OAC-2005572. This work is  supported by the National Institute on Aging of the NIH (Award Number P01AG039347). The funders had no role in study design, data collection and analysis, decision to publish, or preparation of the manuscript.

\bibliography{Reference}

\begin{thebibliography}{19}
\providecommand{\natexlab}[1]{#1}

\bibitem[{Ayoola et~al.(2022)Ayoola, Tyagi, Fisher, Christodoulopoulos, and
  Pierleoni}]{ayoola_refined_2022}
Ayoola, T.; Tyagi, S.; Fisher, J.; Christodoulopoulos, C.; and Pierleoni, A.
  2022.
\newblock ReFinED: An Efficient Zero-shot-capable Approach to End-to-End Entity
  Linking.
\newblock In \emph{Proceedings of the 2022 NAACL: Human Language Technologies:
  Industry Track, {NAACL} 2022}, 209--220.

\bibitem[{Christen(2012)}]{christen_data_2012}
Christen, P. 2012.
\newblock The {Data} {Matching} {Process}.
\newblock In \emph{Data {Matching}: {Concepts} and {Techniques} for {Record}
  {Linkage}, {Entity} {Resolution}, and {Duplicate} {Detection}},
  Data-{Centric} {Systems} and {Applications}, 23--35. Berlin, Heidelberg:
  Springer.
\newblock ISBN 978-3-642-31164-2.

\bibitem[{Das et~al.(2022)Das, Doan, G.~C., Gokhale, Konda, Govind, and
  Paulsen}]{magellandata}
Das, S.; Doan, A.; G.~C., P.~S.; Gokhale, C.; Konda, P.; Govind, Y.; and
  Paulsen, D. 2022.
\newblock The Magellan Data Repository.
\newblock \url{https://sites.google.com/site/anhaidgroup/projects/data}.

\bibitem[{Elmagarmid, Ipeirotis, and
  Verykios(2007)}]{elmagarmid_duplicate_2007}
Elmagarmid, A.~K.; Ipeirotis, P.~G.; and Verykios, V.~S. 2007.
\newblock Duplicate {Record} {Detection}: {A} {Survey}.
\newblock \emph{IEEE Transactions on Knowledge and Data Engineering}, 19(1):
  1--16.
\newblock Conference Name: IEEE Transactions on Knowledge and Data Engineering.

\bibitem[{Enríquez et~al.(2017)Enríquez, Domínguez-Mayo, Escalona, Ross, and
  Staples}]{enriquez_entity_2017}
Enríquez, J.; Domínguez-Mayo, F.; Escalona, M.; Ross, M.; and Staples, G.
  2017.
\newblock Entity reconciliation in big data sources: {A} systematic mapping
  study.
\newblock \emph{Expert Systems with Applications}, 80: 14--27.

\bibitem[{Han et~al.(2021)Han, Zhao, Ding, Liu, and Sun}]{han_2021}
Han, X.; Zhao, W.; Ding, N.; Liu, Z.; and Sun, M. 2021.
\newblock {PTR:} Prompt Tuning with Rules for Text Classification.
\newblock \emph{CoRR}, abs/2105.11259.

\bibitem[{Hulsebos et~al.(2019)Hulsebos, Hu, Bakker, Zgraggen, Satyanarayan,
  Kraska, Demiralp, and Hidalgo}]{hulsebos_sherlock_2019}
Hulsebos, M.; Hu, K.~Z.; Bakker, M.~A.; Zgraggen, E.; Satyanarayan, A.; Kraska,
  T.; Demiralp, {\c{C}}.; and Hidalgo, C.~A. 2019.
\newblock Sherlock: {A} Deep Learning Approach to Semantic Data Type Detection.
\newblock In \emph{Proceedings of the 25th {ACM} {SIGKDD} International
  Conference on {KDD} 2019}, 1500--1508. {ACM}.

\bibitem[{Kojima et~al.(2022)Kojima, Gu, Reid, Matsuo, and
  Iwasawa}]{kojima_large_2022}
Kojima, T.; Gu, S.~S.; Reid, M.; Matsuo, Y.; and Iwasawa, Y. 2022.
\newblock Large Language Models are Zero-Shot Reasoners.
\newblock \emph{CoRR}, abs/2205.11916.

\bibitem[{Li et~al.(2021{\natexlab{a}})Li, Miao, Wang, Sun, and
  Wang}]{li_improving_2021}
Li, B.; Miao, Y.; Wang, Y.; Sun, Y.; and Wang, W. 2021{\natexlab{a}}.
\newblock Improving the {Efficiency} and {Effectiveness} for {BERT}-based
  {Entity} {Resolution}.
\newblock \emph{Proceedings of the AAAI Conference on Artificial Intelligence},
  35(15): 13226--13233.
\newblock Number: 15.

\bibitem[{Li et~al.(2021{\natexlab{b}})Li, Talburt, Li, and
  Liu}]{arabnia_when_2021}
Li, X.; Talburt, J.~R.; Li, T.; and Liu, X. 2021{\natexlab{b}}.
\newblock When Entity Resolution Meets Deep Learning, Is Similarity Measure
  Necessary?
\newblock In \emph{Advances in Artificial Intelligence and Applied Cognitive
  Computing}, 127--140. Springer.

\bibitem[{Li et~al.(2020)Li, Li, Suhara, Doan, and Tan}]{li_deep_2020}
Li, Y.; Li, J.; Suhara, Y.; Doan, A.; and Tan, W.-C. 2020.
\newblock Deep entity matching with pre-trained language models.
\newblock \emph{Proceedings of the VLDB Endowment}, 14(1): 50--60.

\bibitem[{Liu et~al.(2021)Liu, Yuan, Fu, Jiang, Hayashi, and
  Neubig}]{pengfeiliu}
Liu, P.; Yuan, W.; Fu, J.; Jiang, Z.; Hayashi, H.; and Neubig, G. 2021.
\newblock Pre-train, Prompt, and Predict: {A} Systematic Survey of Prompting
  Methods in Natural Language Processing.
\newblock \emph{CoRR}, abs/2107.13586.

\bibitem[{Liu et~al.(2020)Liu, Zhou, Zhao, Wang, Ju, Deng, and
  Wang}]{liu_k-bert_2020}
Liu, W.; Zhou, P.; Zhao, Z.; Wang, Z.; Ju, Q.; Deng, H.; and Wang, P. 2020.
\newblock K-BERT: Enabling Language Representation with Knowledge Graph.
\newblock In \emph{Proceedings of the AAAI}, volume~34, 2901--2908.

\bibitem[{Min et~al.(2021)Min, Ross, Sulem, Veyseh, Nguyen, Sainz, Agirre,
  Heintz, and Roth}]{min_recent_2021}
Min, B.; Ross, H.; Sulem, E.; Veyseh, A. P.~B.; Nguyen, T.~H.; Sainz, O.;
  Agirre, E.; Heintz, I.; and Roth, D. 2021.
\newblock Recent Advances in Natural Language Processing via Large Pre-Trained
  Language Models: {A} Survey.
\newblock \emph{CoRR}, abs/2111.01243.

\bibitem[{Narayan et~al.(2022)Narayan, Chami, Orr, and
  R{\'{e}}}]{narayan_can_2022}
Narayan, A.; Chami, I.; Orr, L.~J.; and R{\'{e}}, C. 2022.
\newblock Can Foundation Models Wrangle Your Data?
\newblock \emph{CoRR}, abs/2205.09911.

\bibitem[{Paganelli et~al.(2022)Paganelli, Buono, Baraldi, and
  Guerra}]{paganelli_analyzing_2022}
Paganelli, M.; Buono, F.~D.; Baraldi, A.; and Guerra, F. 2022.
\newblock Analyzing how {BERT} performs entity matching.
\newblock \emph{Proceedings of the VLDB Endowment}, 15(8): 1726--1738.

\bibitem[{Suhara et~al.(2022)Suhara, Li, Li, Zhang, Demiralp, Chen, and
  Tan}]{suhara_annotating_2022}
Suhara, Y.; Li, J.; Li, Y.; Zhang, D.; Demiralp, c.; Chen, C.; and Tan, W.-C.
  2022.
\newblock Annotating Columns with Pre-Trained Language Models.
\newblock In \emph{Proceedings of the 2022 International Conference on
  Management of Data}, SIGMOD '22, 1493–1503. Association for Computing
  Machinery.

\bibitem[{Trabelsi, Heflin, and Cao(2022)}]{trabelsi_dame_2022}
Trabelsi, M.; Heflin, J.; and Cao, J. 2022.
\newblock {DAME}: {Domain} {Adaptation} for {Matching} {Entities}.
\newblock In \emph{Proceedings of the {Fifteenth} {ACM} {International}
  {Conference} on {Web} {Search} and {Data} {Mining}}, 1016--1024. Virtual
  Event AZ USA: ACM.
\newblock ISBN 978-1-4503-9132-0.

\bibitem[{Vos, D{\"o}hmen, and Schelter(2022)}]{vos2022towards}
Vos, D.; D{\"o}hmen, T.; and Schelter, S. 2022.
\newblock Towards Parameter-Efficient Automation of Data Wrangling Tasks with
  Prefix-Tuning.
\newblock In \emph{NeurIPS 2022 First Table Representation Workshop}.

\end{thebibliography}

\section{APPENDIX}
\paragraph{A. Hyperparameter Settings.} 
Baseline and KAER models share the same hyperparameter setting, in which the batch size equals 64, the max length equals 512, the learning rate is 3e-5, and the number of training epochs is 20. 

\paragraph{B. Baseline models.} 
\begin{table}[!b]
\centering
\begin{tabular}{|l|l|}
\hline
Dataset                  & Injection       \\ \hline
dirty/DBLP-Google        & Ditto + General \\ \hline
dirty/iTunes-Amazon      & Ditto + Product \\ \hline
structured/iTunes-Amazon & Ditto + Product \\ \hline
structured/Amazon-Google & Ditto + Product \\ \hline
structured/DBLP-ACM      & Ditto + General \\ \hline
structured/DBLP-Google   & Ditto + General \\ \hline
textual/Abt-Buy          & Ditto + Product \\ \hline
\end{tabular}
\caption{Datasets and Corresponding Ditto default knowledge injection methods}
\label{tab:dittoinject}
\end{table}
This paper includes two baseline models, i.e., RoBERTa without any knowledge injection and Ditto with its default knowledge injection methods. 

In detail, Table~\ref{tab:dittoinject} shows the datasets and corresponding injection methods utilized by Ditto. "+ General" indicates that Ditto injects seven entity types into the corresponding dataset. The seven types include 'PERSON', 'ORG', 'LOC', 'PRODUCT', 'DATE', 'QUANTITY', and 'TIME'. "+ Product" indicates that Ditto annotates as 'PORODUCT', any entity mentions in the following types, i.e., 'NORP', 'GPE', 'LOC', 'PERSON', and 'PRODUCT'.

\paragraph{C. Two-sided Paired T-test.}
The two-sided paired T-test is implemented to test whether the predictions of different models are significantly different. We first compare whether the predicted label is equal to the ground-truth label. The value is 1 if the prediction is equal to the ground truth and 0 otherwise. And then, the paired T-test is conducted.

\end{document}